%% file: main.tex
\documentclass[10pt,twocolumn,letterpaper]{article}
\usepackage{booktabs,multirow}
\usepackage[table]{xcolor}
\usepackage{tcolorbox}
\usepackage{array}
\usepackage{multirow}
\usepackage{array} % 必须
\usepackage{multirow}
\newcolumntype{C}[1]{>{\centering\arraybackslash}p{#1}}

\newtcolorbox{graylist}{
  colback=gray!5,    
  colframe=gray!15, 
  boxrule=0pt,      
  arc=0mm,          
  left=1mm, right=1mm, top=1mm, bottom=1mm,
  width=\linewidth,          % 用整行宽，更稳
  coltext=black!90, 
  fontupper=\slshape,
  before upper={\raggedright\setlength{\parskip}{2pt}}
}

\usepackage{cvpr}              % To produce the CAMERA-READY version
\input{preamble}

\definecolor{cvprblue}{rgb}{0.21,0.49,0.74}
\usepackage[pagebackref,breaklinks,colorlinks,allcolors=cvprblue]{hyperref}

\def\paperID{41997} % *** Enter the Paper ID here
\def\confName{CVPR}
\def\confYear{2026}

\title{Parse, Search, and Confirmation: Training-Free Aerial Vision-and-Dialog Navigation with Chain-of-Thought Reasoning and Structured Spatial Memory}

\author{
Yu Qi$^{1}$\quad Hongyu Li$^{4}$\quad Shaofei Huang$^{5}$\quad Tianrui Hui$^{1,2,3}$\thanks{Corresponding authors.} \\
Yaxiong Wang$^{1}$\quad Lechao Cheng$^{1}$\quad Zhun Zhong$^{1}$\footnotemark[1]\quad  Si Liu$^{4}$\quad
Meng Wang$^{1}$ \\[2pt]
$^{1}$School of Computer Science and Information Engineering, Hefei University of Technology\\
$^{2}$Jianghuai Advance Technology Center\quad
$^{3}$Anhui Provincial Key Laboratory of Humanoid Robots\\
$^{4}$School of Artificial Intelligence, Beihang University\quad
$^{5}$University of Macau
}
\begin{document}
\maketitle
\input{sec/0_abstract}    
\input{sec/1_intro}
\input{sec/2_related_work}
\input{sec/3_method}
\input{sec/4_experiments}
\input{sec/5_conclusion}
% \section*{Acknowledgements}
% ×××
% \input{supp}
{
    \small
    \bibliographystyle{ieeenat_fullname}
    \bibliography{main}
}


\end{document}

%% file: preamble.tex
\usepackage{enumitem}

%% file: sec/0_abstract.tex
\begin{abstract}
In this paper, we tackle the Aerial Vision-and-Dialog Navigation (AVDN) task in the training-free setting for resource-efficient high-altitude UAV navigation.Naively applying MLLMs leads to unreliable navigation due to weak directional grounding and the lack of explicit spatial memory.
To address these issues, we propose PSC-AVDN, a training-free framework that tightly couples a three-stage Parsing-Search-Confirmation reasoning pipeline with a Structured Spatial Memory (SSM).
The parsing stage uses an LLM to convert ambiguous dialogue instructions into stable geometric directional and destination cues.
A Search Chain-of-Thought (S-CoT) then performs stepwise target exploration under high-altitude observations, and a Confirmation Chain-of-Thought (C-CoT) conducts fine-grained verification around candidate regions to resolve visual ambiguity.
Meanwhile, SSM integrates three complementary sources of spatial cues, including multi-scale visual observation, spatial visual memory, and structured geometric memory to provide global spatial context and long-horizon consistency.
Extensive experiments on ANDH and ANDH-Full show that PSC-AVDN establishes new state-of-the-art performance in the training-free setting, matching or surpassing several finetuned methods.Code will be publicly available at: https://github.com/QY6616/PSC-AVDN
\end{abstract}

%% file: sec/1_intro.tex
\section{Introduction}
\label{sec:Introduction}

\begin{figure}[!t]
    \centering
    \includegraphics[width=\linewidth]{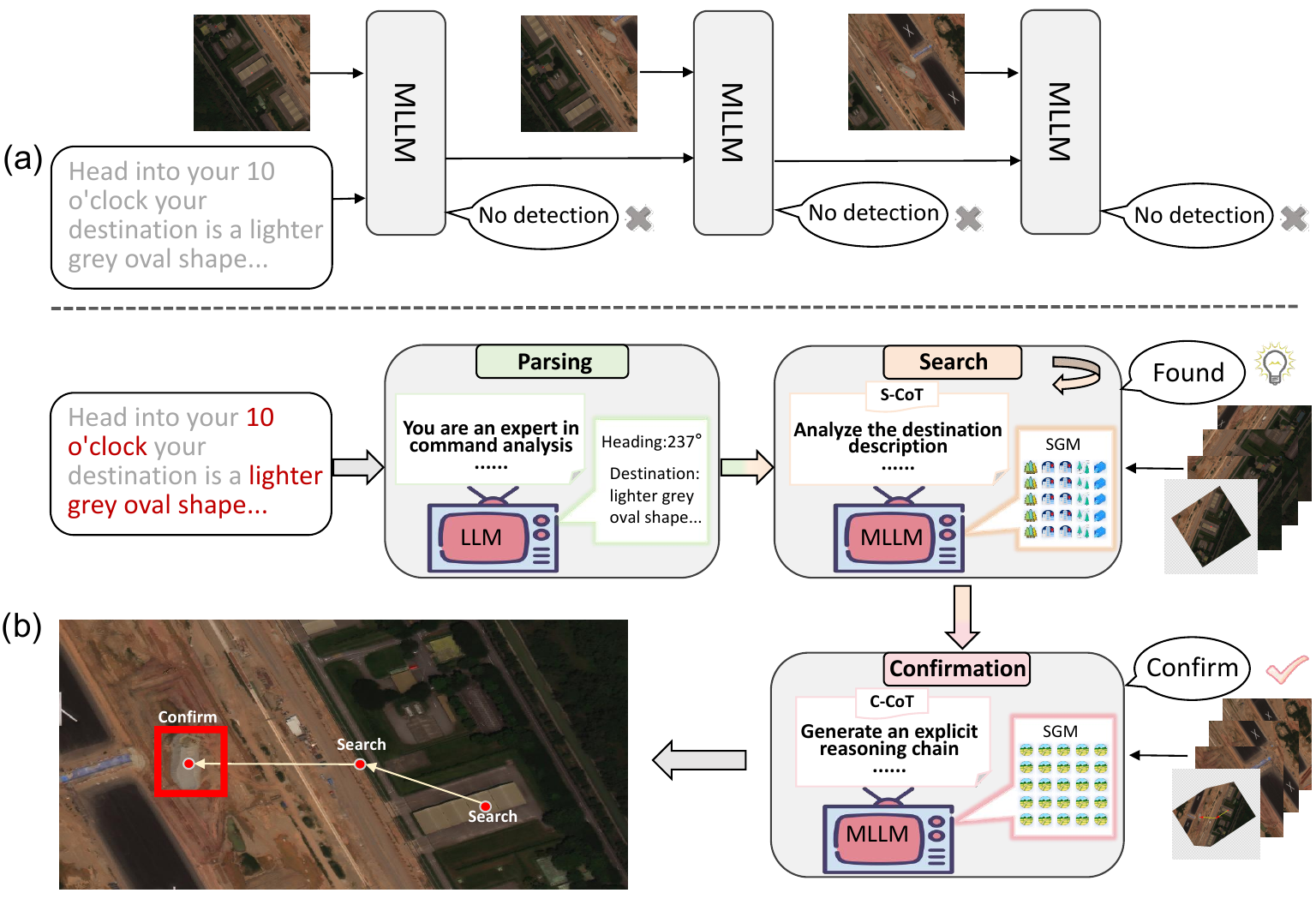}
    \caption{Motivation of our method.
(a) The MLLM baseline suffers from ambiguous directional descriptions and the domain gap between high-altitude imagery and ground-level training data, leading to inaccurate localization.
(b) Our PSC-AVDN eliminates directional ambiguity through instruction parsing, performs structured search via chain-of-thought reasoning, and conducts fine-grained confirmation around the candidate region to achieve more reliable navigation. In addition, a structured spatial memory is introduced to provide clearer spatial context for reasoning.}
    \label{fig:intro}
\end{figure}

Aerial Vision-and-Dialogue Navigation (AVDN)~\cite{Fan22} enables UAVs to follow language instructions while resolving ambiguities through questions and interpreting contextual cues along their flight trajectory.
AVDN adopts a high-altitude, top-down perspective similar to remote sensing imagery.
Compared with low-altitude urban settings~\cite{Lee25,Xiao25,Tian25,Zhao25,Ju2025}, this high-altitude setting covers a broader visual scope where landmarks are small in scale and can be either densely clustered or sparsely distributed, making it challenging to localize and track fine-grained landmarks during navigation.
These characteristics make AVDN well-suited for applications requiring large-scale environmental understanding, such as disaster rescue, environmental monitoring, and geospatial mapping~\cite{Kucharczyk21,Raja24,Ping25,Wang2025}, where operators must reason over wide areas relying on visually subtle landmarks.

In practice, however, AVDN methods rely on supervised finetuning~\cite{Fan22,Su23,Su25}, which entails high computational costs, heavy annotation efforts, and repeated re-annotation and re-training when adapting to new environments.
To overcome these issues, we explore a training-free AVDN framework that enables resource-efficient UAV navigation from high-altitude perspectives, which is particularly difficult without task-specific training.
Thanks to recent advances in Multimodal Large Language Models (MLLMs)~\cite{Li2025,Zhu23,Han24,Luo25,Alibaba23}, we construct a baseline that feeds the UAV’s current view and dialogue instruction into an MLLM and uses task-specific prompts to drive step-by-step target searching until success or timeout.

However, due to the intrinsic mismatch between current MLLMs and the AVDN task requirements, this naive iterative search scheme often yields unreliable navigation. First, MLLMs lack robust spatial grounding and scene understanding under high-altitude perspectives, as their training is dominated by short-range human-view interactions and ground-level imagery, creating a gap between learned visual-linguistic priors and the geometric reasoning needed for aerial navigation. Consequently, MLLMs cannot translate seemingly simple yet abstract expressions, such as \textit{``turn a little to the right''} or \textit{``head toward your 10 o'clock''} (see Figure~\ref{fig:intro}), into meaningful geometric cues reflecting the UAV's spatial layout, leading to early navigation failures. They also struggle to interpret tiny, texture-sparse top-down landmarks in high-altitude scenes, hindering alignment between visual observations and textual instructions and causing the UAV to stop at incorrect locations or miss the target. Second, MLLMs lack global spatial understanding and temporal state tracking, while effective AVDN requires maintaining awareness of visited regions and updating environmental belief over multiple steps. The autoregressive, language-driven inference paradigm offers no structured mechanism for map building or long-range spatial consistency, forcing the model to interpret each view in isolation and resulting in unreliable navigation in complex aerial environments.

To tackle the first limitation, we propose a three-stage \textit{\textbf{Parsing-Search-Confirmation}} framework for training-free AVDN, termed PSC-AVDN, illustrated in Figure~\ref{fig:intro}(b).
The key idea is to decouple directional understanding from high-altitude target localization to make up for MLLM's defect.
The \textbf{\textit{Parsing}} stage converts ambiguous dialogue instructions into reliable directional and destination cues.
The \textbf{\textit{Search}} stage conducts structured target exploration under high-altitude observations.
The \textbf{\textit{Confirmation}} stage performs fine-grained disambiguation around candidate regions to ensure accurate localization.
Concretely, in the \textit{parsing} stage, we employ a general-purpose LLM~\cite{Liu24} to extract directional and destination cues from the dialogue instruction and translate multiple directional formats into a consistent angular representation.
A Heading Resolution (HR) module is designed to produce a stable and executable direction signal for subsequent navigation.
For example, in Figure~\ref{fig:intro}(b), the abstract \textit{`$10$ o'clock'} orientation is converted into an absolute \textit{`$237^\circ$'} in the parsing stage.
In the \textit{search} stage, we design a Search Chain-of-Thought (S-CoT) mechanism that decomposes target search into a sequence of interpretable reasoning steps.
This structured reasoning progressively narrows down candidate regions.
However, in high-altitude scenarios, the target remains small and often ambiguous even near the correct region.
To resolve visual ambiguity, we introduce the final \textit{confirmation} stage, where a Confirmation Chain-of-Thought (C-CoT) performs fine-grained verification around the candidate region identified by the \textit{search} stage.
By checking spatial and relational constraints against the destination description, C-CoT disambiguates tiny and texture-sparse aerial targets and confirms the unique location that satisfies all instruction-derived conditions.

To tackle the second limitation, we introduce a Structured Spatial Memory (SSM) module to provide global spatial context and history information that MLLMs lack to enhance the search-confirmation process.
SSM supplies spatial cues.
(i) Multi-scale visual observation, obtained from global, local, and main-view crops to form a hierarchical perception of large-scale scene layout and fine-grained geometric details.
(ii) Spatial visual memory, constructed by fusing historical crops with the current view to form a persistent trajectory representation that maintains awareness of visited regions and reduces long-horizon drift.
(iii) Structured geometric memory, built by an updated reference grid map whose cells store semantic categories inferred over time, providing a structured prior that stabilizes multi-step spatial reasoning.
Leveraging these cues, SSM improves navigation stability in large-scale aerial environments.

The main contributions of this work are summarized as follows:
(1) We propose PSC-AVDN, the first training-free framework for Aerial Vision-and-Dialog Navigation task that designs structured reasoning strategies for instruction Parsing, hierarchical target Search, and fine-grained Confirmation to achieve accurate navigation under complex high-altitude scenarios.
(2) We propose a Structured Spatial Memory (SSM) module that explicitly provides global spatial context and history information to enhance the search-confirmation process.
(3) Extensive experimental results demonstrate that our method achieves state-of-the-art performance under the training-free setting on both ANDH and ANDH-Full datasets, comparable to or even surpassing several finetuning-based methods.

%% file: sec/2_related_work.tex
\section{Related Work}
\label{sec:Related Work}
\subsection{Vision-and-Language Navigation}
Vision-and-Language Navigation (VLN)~\cite{Anderson18,Krantz20,Ku20,Moudgil21,Ju2026} trains an agent to follow natural language instructions to reach a target in indoor panoramic environments, driving advances in modality alignment and policy learning.
VLN-CE~\cite{Krantz20} increases task complexity by requiring continuous-space navigation using low-level actions without prior maps or localization.
To overcome limited data and domain shift, later works employ large-scale pretraining.
For example, Lin \textit{et al.}~\cite{Lin23} use YouTube house tour videos to build path-instruction pairs that greatly enhance navigation performance.
More recent studies explore training-free paradigms.
DiscussNav-GPT4~\cite{Long24} coordinates multiple expert models in a multi-agent discussion framework, while Open-Nav~\cite{Qiao25} introduces a spatiotemporal chain-of-thought reasoning process comprising instruction comprehension, progress estimation, and action execution, improving LLM-based perception and reasoning in navigation.

\subsection{Aerial Vision-and-Language Navigation}
Aerial VLN~\cite{Lin23,Gao24,Wang25b} (AVLN) extends the VLN paradigm to aerial environments, enabling UAVs to navigate complex landscapes via natural language instructions and real-time visual inputs. OpenUAV~\cite{Wang25b} provides a realistic UAV simulator with 12,000 trajectories and the ``UAV-Need-Help'' benchmark, while CityNav~\cite{Lee25} offers a large-scale urban dataset spanning 4.65 km$^2$ with 32,637 human trajectories. Building on these resources, NavAgent~\cite{Liu24b} and OpenFly~\cite{Gao25b} enhance language grounding and multimodal perception through multi-scale fusion, graph encoding, and keyframe-aware policies.  
Unlike prior AVLN studies focusing on city-scale environments and egocentric views, high-altitude UAV navigation remains underexplored. AVDN~\cite{Fan22} addresses this setting, where observations cover vast areas with small, weak landmarks. However, existing AVDN methods~\cite{Su23,Su25} rely on supervised training, incurring high annotation costs and limited cross-domain generalization. To our knowledge, we are the first to investigate a training-free AVDN framework based on three-stage structured reasoning.

\subsection{Spatial Reasoning with MLLMs}
Recent studies~\cite{Shao24,Hu24,Chen24,Cheng24,yu2026open,Huang2025} indicate that the reasoning paradigm of MLLMs is shifting from purely language-based chain-of-thought reasoning to approaches incorporating visual thinking and spatial memory.
MVoT~\cite{Li25} introduces visualized reasoning trajectories that alternate with text, reducing information loss in dynamic and complex scenes.
VSI-Bench~\cite{Yang25} evaluates visual-spatial intelligence from videos, showing that language-centric heuristics contribute little to spatial tasks, while cognitive map construction improves distance estimation and layout understanding.
TopV-Nav~\cite{Zhong24} and MapNav~\cite{Zhang25} further enhance spatial memory by integrating annotated textual cues into semantic map representations.
In the domain of UAV navigation, GeoNav \cite{Xu25} adopts a cognitive map paradigm based on manual annotations.
However, its construction of cognitive maps relies on external models such as Grounded SAM~\cite{Ren24} and OpenGIS priors, which increases deployment complexity and cost.
In contrast, our approach is entirely built upon the native visual understanding capabilities of MLLMs~\cite{Dong25,Schulze25}.
We prompt the model within the reasoning chain to generate a reference grid map and incorporate both historical trajectories and multi-scale views, thereby providing enriched structured spatial memory without any external dependencies.

%% file: sec/3_method.tex
\section{Method}
\label{sec:method}

\begin{figure*}[!t]
    \centering
    \includegraphics[width=\linewidth]{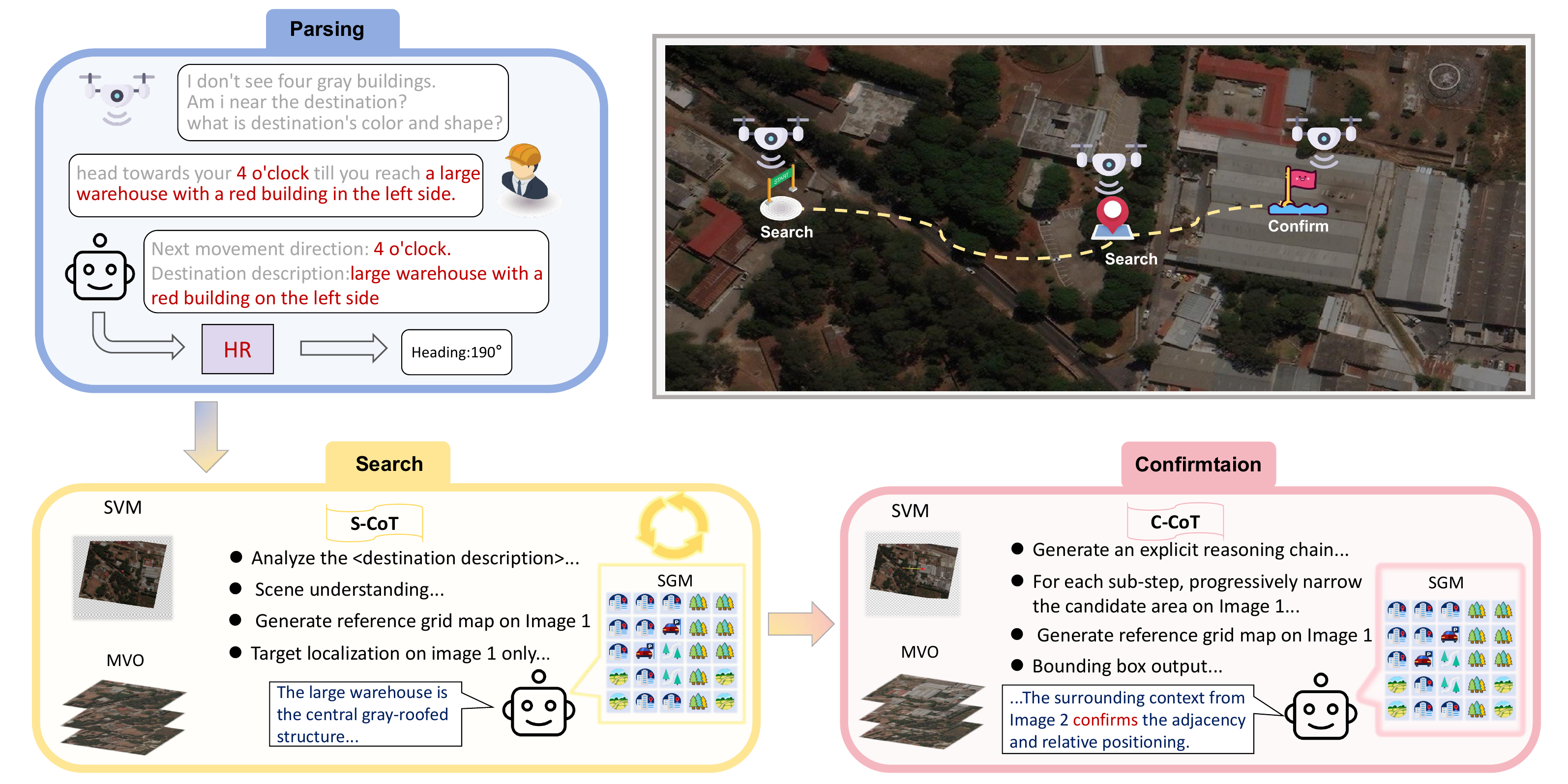}
    \caption{The overall architecture of our proposed Parsing-Search-Confirmation framework for Aerial Vision-and-Dialog Navigation (PSC-AVDN). (a) The three-stage reasoning process first parses the destination and direction, followed by navigation through the step-wise reasoning chain (S-CoT and C-CoT), gradually searching and confirming the target location. (b) The Structured Spatial Memory (SSM) module provides multi-scale visual observation (MVO), spatial visual memory (SVM), and structured geometric memory (SGM) to enhance the search-confirmation process.}
    \label{fig:2}
\end{figure*}

\subsection{Overview}
As shown in Figure~\ref{fig:2}, our PSC-AVDN framework performs navigation through a three-stage parsing-search-confirmation (PSC) reasoning pipeline enhanced by a Structured Spatial Memory (SSM) to provide historical-aware clues.
The AVDN task consists of $L$ dialogue rounds, where in each round $l$, our PSC-AVDN framework receives an text instruction $\mathcal{U}_l$ and executes a complete PSC cycle to localize and move to the corresponding target $\mathcal{B}_l$, which is denoted as a bounding box $\mathcal{B}_l=(x_l^1,y_l^1,x_l^2,y_l^2)$.
The last bounding box $\mathcal{B}_L$ is regarded as the final target of the navigation process.
We formalize each round as a cross-modal mapping:
\begin{equation}
    \mathcal{F} : (\mathcal{U}_l, \{\mathcal{V}_t\}_{t=t_l^s}^{t_l^e}, \mathcal{M}_{t_l^e}, \mathcal{R}_{t_l^e}) \rightarrow \mathcal{B}_l,
\end{equation}
where $\mathcal{V}_t$, $\mathcal{M}_t$, and $\mathcal{R}_t$ denote the multi-scale visual observation, spatial visual memory and structured geometric memory at step $t$ generated by our SSM module.
$t_l^s$ and $t_l^e$ represent the start and end time steps in the $l$-th round.

For the $l$-th round of dialogue with $l\in [1, L]$, PSC-AVDN first parses the instruction $\mathcal{U}_l$ into an explicit destination cues and an absolute angle indicating the flying direction of UAV in the parsing stage.
These initial cues guide a unified CoT-driven reasoning process, where $\mathcal{V}_t$, $\mathcal{M}_t$, $\mathcal{R}_t$ jointly support coarse localization in the search stage and the subsequent semantic disambiguation in the confirmation stage, yielding the predicted target region $\mathcal{B}_l$.
After each step of the above process, the spatial memories are updated in the SSM module as:
\begin{equation}
\mathcal{M}_{t+1}, \mathcal{R}_{t+1}=\mathrm{SSM}(\mathcal{M}_{t}, \mathcal{R}_t).
\end{equation}
For clarity of exposition, the subsequent method description focuses on a single dialogue round, with the index $l$ omitted when no ambiguity arises.

\subsection{Parsing-Search-Confirmation Reasoning}
\subsubsection{Parsing Stage}
We first employ a general-purpose LLM to structure and decompose the dialogue instructions 
$\mathcal{U}$, during which the movement direction phrase $s_{\text{dir}}$ and the destination description $s_{\text{des}}$ are extracted.
Since $s_{\text{dir}}$ may appear in various textual forms (\textit{e.g.}, ``$3$ o'clock'', ``$120^\circ$'', or ``north-east''), we design a Heading Resolution (HR) module to convert these heterogeneous expressions into a canonical angular representation.
Formally, HR maps the textual direction $s_{\text{dir}}$ to an absolute angle $\alpha$ through a rule-based parser that transforms clock-based expressions, degree-marked notations, and compass directions into angles within \([0,2\pi)\). 
Based on the UAV's current azimuth $\phi$, HR further computes the relative heading:
\begin{equation}
\delta = \mathrm{wrap}\!\left(\alpha - \phi\right),
\end{equation}
where $\mathrm{wrap}(\cdot)$ denotes normalizing the angle to the standard range $[0, 2\pi)$.
Through the above process, the resulting relative heading $\delta$ is then combined with the destination description $s_{\text{des}}$ and fed into the subsequent reasoning stages.

\subsubsection{Search Stage}
In this stage, we design a Search Chain-of-Thought (S-CoT) to guide the MLLM through step-wise reasoning that progressively identifies candidate regions. S-CoT divides this process into four sequential sub-reasoning steps:

\begin{itemize}[leftmargin=*]
\item \textbf{Destination analysis,} where the MLLM performs semantic analysis of the destination description $s_{\text{des}}$ to form a task-relevant prior for subsequent reasoning.
Using the instruction \textit{``large warehouse with a red building on the left side''} in Figure~\ref{fig:2} as an example, the model extracts core attributes relevant to visual localization, including the target category (\textit{``warehouse''}), salient reference cues (\textit{``red building''}), and the spatial relation (\textit{``on the left side''}).
These attributes jointly constitute a structured description of the target's visual characteristics and geometric relations as explicit constraints.
\item \textbf{Scene understanding,} where the MLLM is commanded to construct a holistic understanding of the current view using multi-scale visual observation $\mathcal{V}_t$ and spatial visual memory $\mathcal{M}_t$ at step $t$ generated by our SSM module. (corresponding to MVO and SVM in Figure~\ref{fig:2}).
\item \textbf{Reference grid map generation,} where the MLLM is prompted to divide the main view into a $N \times N$ grid and assign labels from a predefined set of categories to each grid cell, generating the structured geometric memory $\mathcal{R}_t$.
The generation process of $\mathcal{R}_t$ can aid the model to comprehend the visual scene more structurally.
\item \textbf{Target localization,} where the model identifies candidate target regions in the main view based on visual features and destination information.
Once a candidate region is successfully located, our pipeline proceeds to the final confirmation stage.
\end{itemize}
Through this explicit step-by-step reasoning mechanism, S-CoT progressively narrows down the candidate areas, enhancing the stability and interpretability of the MLLM in the complex aerial target search task.

\subsubsection{Confirmation Stage}
After the search stage, the UAV can reach the vicinity of the target area. However, in the AVDN task, the high-altitude viewpoint results in small target sizes, large scale variations, weak landmark distinctiveness, and complex spatial-semantic correlations with the surrounding environment.
These task-specific challenges make it difficult for the model to achieve precise localization relying solely on S-CoT.
Therefore, we introduce the Confirmation Chain-of-Thought (C-CoT) to address this issue.
The core idea of C-CoT is to refine and disambiguate candidate regions through interpretable reasoning chains.
First, the model is forced to generate verifiable step-by-step reasoning based on the destination description.
It then verifies spatial and relational constraints using multiple-scale views, gradually excluding incorrect candidates based on finer local structures, directions, and adjacency relationships.
Ultimately, the model confirms the unique target location and outputs the confidence level along with a brief visual evidence explanation.
For instance, when the instruction is \textit{``large warehouse with a red building on the left side''}, the model first identifies the large warehouse, then checks whether a red building exists on its left side, and finally verifies the spatial adjacency relationship between the two.
Finally, the model determines the unique target region and outputs the bounding box for current round $\mathcal{B}$.
It is important to note that in C-CoT, we also prompt the model to generate a reference grid map to assist spatial perception, which is elaborated in the next section.

\begin{figure}[!htbp]
    \centering
    \includegraphics[width=\linewidth]{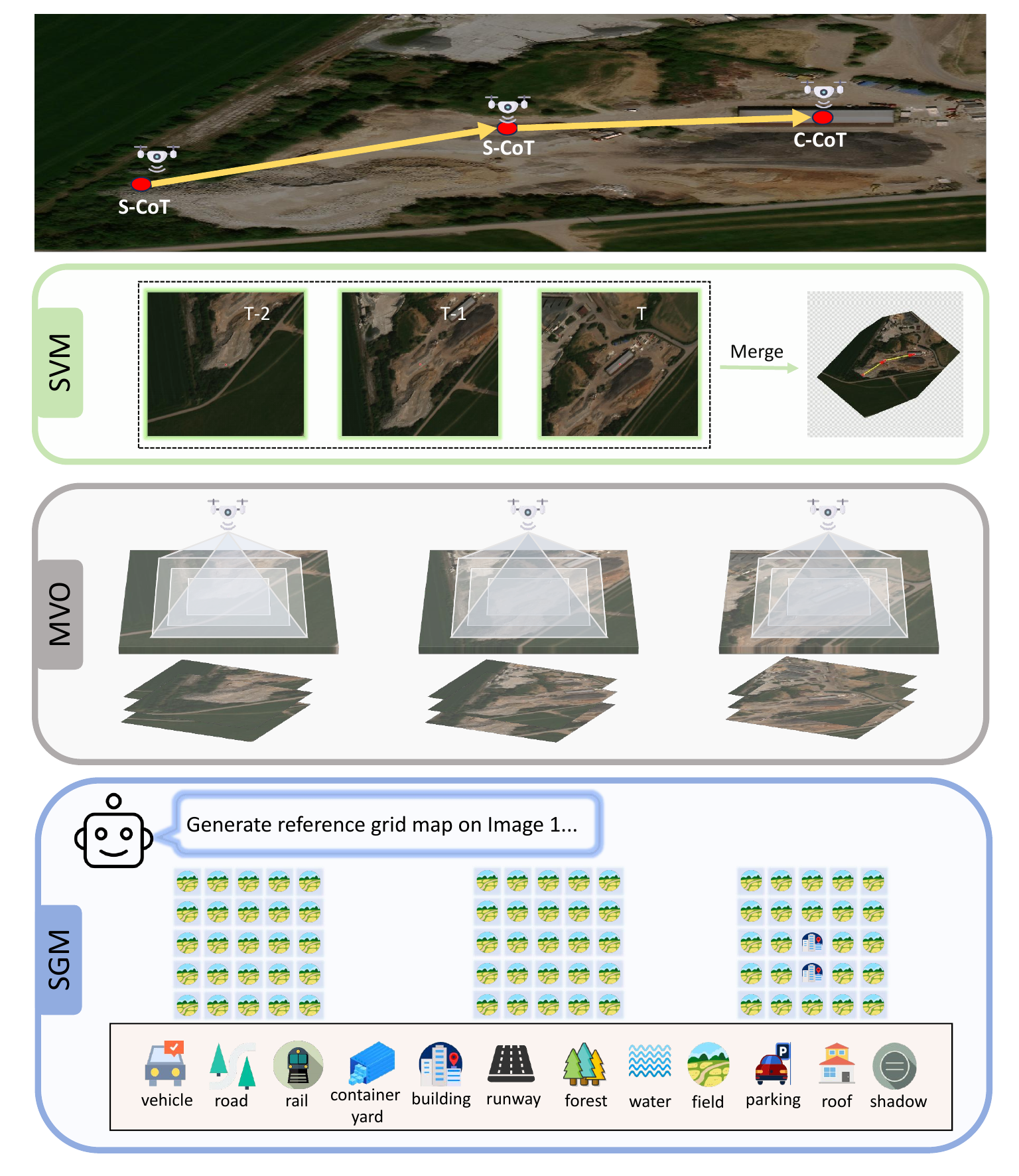}
    \caption{SSM Module diagram. A concrete case is presented to demonstrate how the SSM module operates within the CoT process. The module consists of three main parts: (a) Multi-scale Visual Observation (MVO): Visual inputs at different scales help the model acquire various levels of visual information. (b) Spatial Visual Memory (SVM): Historical information is fused with the current CoT to ensure temporal and spatial continuity and consistency in the reasoning process. (c) Structured Geometric Memory (SGM): The model is guided to generate a reference grid map to assist in spatial perception and reasoning.}
    \label{fig:3}
\end{figure}

\begin{table*}[!htbp]
\centering
\setlength{\tabcolsep}{3.2pt}
\renewcommand{\arraystretch}{1.15}
\definecolor{tfpink}{RGB}{255,235,240}
\definecolor{tfpink2}{RGB}{255,220,228}

{\setlength{\aboverulesep}{0pt}%
 \setlength{\belowrulesep}{0pt}%
 \resizebox{\textwidth}{!}{%

% -------------------------
% 列格式修改：ANDH 与 ANDH-Full 之间加入竖线 |
% 小组内部保持 ccc（无竖线）
% -------------------------
\begin{tabular}{@{\hspace{3.3pt}} l|
ccc|ccc|ccc
|ccc|ccc|ccc@{\hspace{3.3pt}}}
\toprule

\rowcolor{gray!10}
\multicolumn{1}{c|}{\cellcolor{gray!10}}
& \multicolumn{9}{c|}{\textbf{ANDH}}
& \multicolumn{9}{c}{\textbf{ANDH-Full}} \\

\rowcolor{gray!10}
\multicolumn{1}{c|}{\cellcolor{gray!10}\textbf{Method}}
& \multicolumn{3}{c}{Seen Val.} 
& \multicolumn{3}{c}{Unseen Val.} 
& \multicolumn{3}{c|}{Unseen Test}
& \multicolumn{3}{c}{Seen Val.} 
& \multicolumn{3}{c}{Unseen Val.} 
& \multicolumn{3}{c}{Unseen Test} \\

\rowcolor{gray!10}
\multicolumn{1}{c|}{\cellcolor{gray!10}}
& \multicolumn{3}{c}{\textbf{SPL~~~~SR~~~~GP}}
& \multicolumn{3}{c}{\textbf{SPL~~~~SR~~~~GP}}
& \textbf{SPL} & \textbf{SR} & \textbf{GP}
& \multicolumn{3}{c}{\textbf{SPL~~~~SR~~~~GP}}
& \multicolumn{3}{c}{\textbf{SPL~~~~SR~~~~GP}}
& \textbf{SPL} & \textbf{SR} & \textbf{GP} \\
\midrule

\multicolumn{19}{c}{\textbf{Supervised Finetuning Methods}}\\
\midrule

E.T.~\cite{Pashevich21}                        
&12.1 & 14.1 & 50.1  &14.3 & 16.6 & 51.9  &11.3 & 13.3 & 51.7
& 2.2 &  3.1 & 51.3 & 2.5 & 3.7 & 48.9 & 1.9 & 2.8 & 60.7 \\

HAA-T~\cite{Fan22}                       
&14.7 &17.3 &56.3 &16.5 &20.4 &55.2 &12.9 &15.7 &53.7
&3.7 &5.1 &54.6 &3.2 &4.7 &50.9 &4.1 &6.3 &63.2 \\

LSTM~\cite{Greff16}                        
& 9.0 &10.3 &31.9 &13.3 &14.1 &35.9 & 9.7 &10.8 &40.4
&1.0 &1.0 &43.8 &3.2 &3.7 &48.7 &1.8 &1.9 &56.4 \\

HAA-LSTM~\cite{Fan22}                    
&11.6 &13.0 &50.3 &18.3 &20.0 &54.4 &12.6 &14.1 &54.6
&3.8 &4.1 &52.2 &3.4 &3.7 &56.1 &1.9 &2.6 &66.5 \\

TA-GAT~\cite{Su23}                      
&12.9 &16.0 &56.9 &18.8 &23.3 &54.3 &15.1 &18.7 &56.5
&\textemdash &\textemdash &\textemdash &\textemdash &\textemdash &\textemdash &\textemdash &\textemdash &\textemdash \\

TA-GAT w/ at~\cite{Su23}                
&14.8 &18.2 &58.8 &17.8 &21.1 &61.7 &15.9 &19.7 &56.3
&\textemdash &\textemdash &\textemdash &\textemdash &\textemdash &\textemdash &\textemdash &\textemdash &\textemdash \\

FELA~\cite{Su25}                        
&15.1 &18.8 &\underline{60.8} &17.2 &20.6 &63.0 &16.4 &20.3 &56.7
&\textemdash &\textemdash &\textemdash &\textemdash &\textemdash &\textemdash &\textemdash &\textemdash &\textemdash \\

FELA w/ at~\cite{Su25}                  
&\underline{15.3} &18.8 &60.7 &\underline{19.2} &\underline{23.9} &\underline{64.1} &\underline{17.6} &\underline{21.9} &\underline{61.4}
&\textemdash &\textemdash &\textemdash &\textemdash &\textemdash &\textemdash &\textemdash &\textemdash &\textemdash \\

OpenFly~\cite{Gao25b} 
& 14.1 & \underline{21.9} & 44.5
& 16.4 & 20.1 & 47.2
& 16.2 & 15.2 & 52.2
& \underline{14.0} & \underline{10.1} & \underline{61.0}
& \underline{12.4} & \underline{15.0} & \underline{59.0}
& \underline{8.3}  & \underline{13.3} & \underline{54.3} \\
\midrule

\multicolumn{19}{c}{\textbf{Training-Free Methods}}\\
\midrule

\rowcolor{tfpink}
GPT-4o~\cite{Shahriar24}                      
& 2.6 & 2.7 & -9.5 & 3.4 & 3.9 & -11.8 & 2.6 & 2.9 & -15.5
& 2.5 & 2.5 & -5.7 & 2.8 & 3.3 & -15.0 & 1.8 & 1.9 & -10.2 \\

\rowcolor{tfpink}
Qwen-VL-Max~\cite{Alibaba23}                 
& 8.5 & 8.6 & 3.6 & 8.7 & 9.2 & 5.5 & 5.7 & 6.2 & 6.7
& 9.3 & 9.6 & 5.2 & 6.8 & 7.0 & 9.1 & 6.4 & 6.9 & 1.6 \\

\rowcolor{tfpink2}
\textbf{PSC-AVDN (Ours)}
& \textbf{16.3} & \textbf{18.6} & \textbf{37.4}
& \textbf{17.8} & \textbf{22.6} & \textbf{39.2}
& \textbf{13.5} & \textbf{16.4} & \textbf{28.2}
& \textbf{19.1} & \textbf{22.3} & \textbf{75.1}
& \textbf{12.4} & \textbf{15.4} & \textbf{62.3}
& \textbf{12.1} & \textbf{14.4} & \textbf{54.5} \\

\bottomrule
\end{tabular}%
}}
\caption{Comparison results on the ANDH and ANDH-Full datasets. Higher values indicate better performance. \underline{Underline} and \textbf{bold} indicate the best results among supervised finetuning and training-free methods, respectively. Our PSC-AVDN achieves state-of-the-art performance in the training-free setting, comparable to or even surpassing several supervised finetuning methods.}
\label{tab:andh_results}
\end{table*}

\subsection{Structured Spatial Memory}
\label{sec:ssm}
To address the limitation caused by relying solely on single-frame main-view images, we introduce the Structured Spatial Memory (SSM) module (Figure~\ref{fig:3}). This module explicitly supplements multi-level spatial and historical information during reasoning, compensating for the inherent spatiotemporal shortcomings of MLLMs and enhancing the stability and reliability of the search and confirmation process. The SSM consists of three core components that progressively operate within the reasoning chain, providing complementary spatial cues for the model.

% \textbf{Multi-Scale Crop.}
% Multi-scale cropping provides visual inputs at different resolutions, allowing the model to perceive scenes from global to local perspectives. For each scale $i$, the image slice $\mathcal{V}_i$ represents the visual information at that scale, formally defined as:
% \begin{equation}
% \mathcal{V}_i = \mathcal{T}_i(\mathcal{I}, s_i), \ \ \text{where} \ \ \mathcal{T}_i(\mathcal{I}, s_i) = {\rm{Resample}}(\mathcal{I}, s_i).
% \end{equation}
% Here, $\mathcal{I}$ is the original image, and $s_i$ is the scaling factor.
% $\mathcal{T}_i(\mathcal{I}, s_i)$ represents the operation of resampling the original image $\mathcal{I}$ at the scaling factor $s_i$ to obtain the cropped image at that scale.
% The final multi-scale images are concatenated as a unified visual view input $\mathcal{V}_{\rm{final}} = \left[ \mathcal{V}_1, \mathcal{V}_2, \dots, \mathcal{V}_m \right]$, where $m$ denotes the number of scales.
\textbf{Multi-scale visual observation.}
Multi-scale visual observation provides image slices captured at different \emph{perceptual scales}, enabling the model to build a hierarchical understanding of both large scale scene layouts and fine grained geometric details from global to local views.
For each scale $i$ at the $t$-th step, the image crop $\mathcal{V}_t^i$ represents the visual information of the corresponding scale:
\begin{equation}
\mathcal{V}_t^i = \mathrm{Resample}(\mathcal{I}, s^i).
\end{equation}
Here, $\mathcal{I}$ denotes the global remote sensing imagery, and $s^i$ is the scaling factor corresponding to the perceptual scale.
% The operator $\mathcal{T}_i(\mathcal{I}, s_i)$ resamples the original image at scale $s_i$ to generate the cropped view at that level.
The final multi-scale visual observation is obtained by combining image crops of all scales:
\begin{equation}
\mathcal{V}_t = 
\left[ \mathcal{V}_t^1, \mathcal{V}_t^2, \dots, \mathcal{V}_t^M \right],
\end{equation}
where $M$ denotes the number of total perceptual scales.

\textbf{Spatial visual memory.}
Spatial visual memory fuses the main view from previous steps together with the UAV's trajectory and orientation information, constructing a coherent and consistent spatial visual memory map. A fixed-size memory canvas is first initialized in a global map coordinate system.
At each time step $t$, the spatial visual memory map $\mathcal{M}_t$ is updated by concatenating the previous time step's memory map $\mathcal{M}_{t-1}$ with the middle scale of $\mathcal{V}_t$ of current view, which is denoted as $\mathcal{V}_t^m$.The concatenated map are then projected onto the memory canvas, where a unified spatial mask is generated via a convex hull operation to represent the accumulated coverage.
In addition, to enhance the stability of spatial memory and ensure spatiotemporal consistency, historical orientation $\theta_t$ and trajectory information $\mathcal{T}_t$ are also fused with the concatenated memory map as:
\begin{equation}
    \mathcal{M}_t = \left( \mathcal{M}_{t-1} \oplus \mathcal{V}_t \right) \oplus \left( \mathcal{T}_t \oplus \theta_t \right)
\end{equation}
where $\oplus$ denotes the concatenation operation.
Spatial visual memory fuses historical crops with the current view to construct a continuous and robust trajectory representation, thereby maintaining awareness of previously explored regions and effectively suppressing long-range drift.

\textbf{Structured geometric memory.}
Structured geometric memory is implemented by prompting the model to generate a reference grid map. The reference grid map divides the main view into discrete grid cells, assigning semantic labels to each cell to provide a structured representation of spatial locations. We statistically derive 12 high-frequency semantic categories from the dataset dialogue prompt to cover most scenarios
Given the middle scale of the current view $\mathcal{V}_t^m$, we partition it into an $N \times N$ grids equally along height and width dimensions, and assign a semantic label $c_j \in \mathcal{C}, j\in [1, N^2]$ to the $j$-th grid, where $\mathcal{C}$ is the set of all possible semantic labels.
The reference grid map for the current step is obtained as:
\begin{equation}
    \mathcal{\bar{R}}_t = \left[ r_1, r_2, \dots, r_{N^2} \right], \ \ \text{where} \ \ r_j = \left( p_j, c_j \right).
\end{equation}
where $p_j$ denotes the spatial coordinate of grid $j$ and $c_j$ denotes its semantic label.
Then the structured geometric memory for time step $t$ is updated as:
\begin{equation}
\mathcal{R}_t = \rm{Update}(\mathcal{R}_{t-1}, \mathcal{\bar{R}}_t).
\end{equation}
Structured geometric memory provides stable support for multi-step spatial reasoning by continuously updating the semantic information within grid cells.

% Together, these components allow the SSM module to incorporate multi-level spatial priors into the reasoning process. Multi-scale observation provides hierarchical scene cues, spatial memory preserves continuity, and geometric memory offers a stable grid-based abstraction. Their combined effect strengthens the model’s understanding of aerial-view environments and improves multi-step reasoning and navigation robustness.

%% file: sec/4_experiments.tex
% \usepackage{booktabs}
% \usepackage{multirow}
% \usepackage{makecell}
% \usepackage[table]{xcolor}
% \usepackage{graphicx}
% \usepackage[table]{xcolor}
% \usepackage{booktabs,multirow}
\section{Experiments}
\label{sec:Experiments}

\subsection{Datasets and Evaluation Metrics}
We conduct experiments on two datasets, namely ANDH~\cite{Fan22} and ANDH-Full~\cite{Fan22}.
The ANDH dataset contains $370$ sub-trajectories in the seen validation set, $411$ sub-trajectories in the unseen validation set, and $897$ sub-trajectories in the unseen test set. In contrast, the ANDH-Full dataset comprises $197$ full trajectories in the seen validation set, $214$ full trajectories in the unseen validation set, and $432$ full trajectories in the unseen test set.

Following prior works~\cite{Fan22,Su23,Su25}, we adopt three evaluation metrics, including Success Rate (SR), Success weighted by Path Length (SPL), and Goal Progress (GP).
SR measures whether the agent reaches the goal region.  SR measures the proportion of successful trajectories. SPL evaluates path efficiency by weighting success with trajectory length. GP quantifies how much progress the agent makes toward the goal by comparing the distance traveled with the remaining distance to the goal.

\subsection{Implementation Details}
For the parsing stage, we adopt the DeepSeek-V3~\cite{Liu24} model to interpret natural language instructions. 
Then, the Qwen-VL-Max~\cite{Alibaba23} model is employed to perform a Search and Confirmation CoT reasoning process, with the maximum execution steps set to $3$. 
A $5\times5$ reference grid is maintained, and historical trajectories are projected onto the base map with a north-up absolute orientation to ensure spatial alignment across steps. 
Our method performs reasoning under multiple visual scales, using scale factors of $3$, $5$, and $7$. 
All visual patches are generated by applying a homography that warps the real-world quadrilateral field-of-view into a canonical $768\times768$ pixel frame, preserving a consistent mapping between pixel coordinates and geographic locations. 
Specifically, we first compute the pixel positions of the four FOV corners on the basemap according to their latitude and longitude, and then estimate the perspective transformation matrix between these points and a standard square, enabling the inverse mapping of any patch pixel back to its corresponding geographic coordinates.

\subsection{Comparison with State-of-the-Art Methods}
Table~\ref{tab:andh_results} presents the quantitative comparison results between our method and previous state-of-the-art approaches on the ANDH and ANDH-Full datasets.
On the ANDH dataset, our proposed training-free PSC-AVDN method achieves performance comparable to or even surpassing several supervised finetuning methods on the three evaluation metrics, yielding the best results in the training-free setting as well.
On the ANDH-Full dataset with longer trajectories, our method also attains overall state-of-the-art performance.
These results show that by designing effective chain-of-thought reasoning and providing structured spatiotemporal memories, we can fully unleash the visual perception and reasoning capabilities of MLLMs, enabling an effective training-free pipeline for the AVDN task. Our method consistently improves performance across different MLLMs (see Appendix 5).

\newcolumntype{C}[1]{>{\centering\arraybackslash}p{#1}}

\begin{table}[!htbp]
\centering
\small
\setlength{\tabcolsep}{1.8pt}
\renewcommand{\arraystretch}{1.10}

{\setlength{\aboverulesep}{0pt}%
 \setlength{\belowrulesep}{0pt}%
\begin{tabular}{@{\hspace{1.8pt}} c c c | C{1.0cm} C{1.0cm} C{1.0cm} @{\hspace{1.8pt}}}
 \toprule
 \rowcolor{gray!10}
 \textbf{Parsing} & \textbf{Search} & \textbf{Confirmation} & \textbf{SPL} & \textbf{SR} & \textbf{GP} \\
 \midrule
   &   &   &  8.7 &  9.2 &  5.5 \\
 \checkmark &   &   & 13.5 & 14.6 & 26.2 \\
 \checkmark & \checkmark &   & 15.6 & 17.5 & 25.8 \\
 \checkmark & \checkmark & \checkmark & \textbf{16.3} & \textbf{19.3} & \textbf{35.7} \\
 \bottomrule
 \end{tabular}%
}
\caption{Ablation results of our three-stage reasoning framework.}
\label{tab:val_unseen_results}
\end{table}

\subsection{Ablation Studies}
We conduct ablation studies on the \textit{Unseen Val.} set of the ANDH dataset to evaluate the effectiveness of different components in our method.
\newcolumntype{C}[1]{>{\centering\arraybackslash}p{#1}}

\textbf{Three-stage reasoning.}
Table~\ref{tab:val_unseen_results} shows the ablation results of our three-stage reasoning framework.
The first row denotes the baseline method using Qwen-VL-Max, which takes simple instruction prompts and current visual observations as input, and performs multi-step iterative searches to locate the target region.
The second row incorporates the Parsing Stage, enabling the model to explicitly extract two types of key information, namely direction and destination description.
This structured extraction provides semantic priors that significantly improve navigation performance.
Building on this, adding the Search Stage further improves performance.
This verifies that the S-CoT design enhances the MLLM's spatial awareness and temporal-spatial reasoning.
Finally, the Confirmation Stage enables fine-grained verification and disambiguation of target-adjacent regions through interpretable reasoning chains.
Overall, the results indicate that each stage progressively improves model performance, confirming the effectiveness of our three-stage reasoning framework and chain-of-thought design.

\begin{figure*}[!h]
    \centering
    \includegraphics[width=\linewidth]{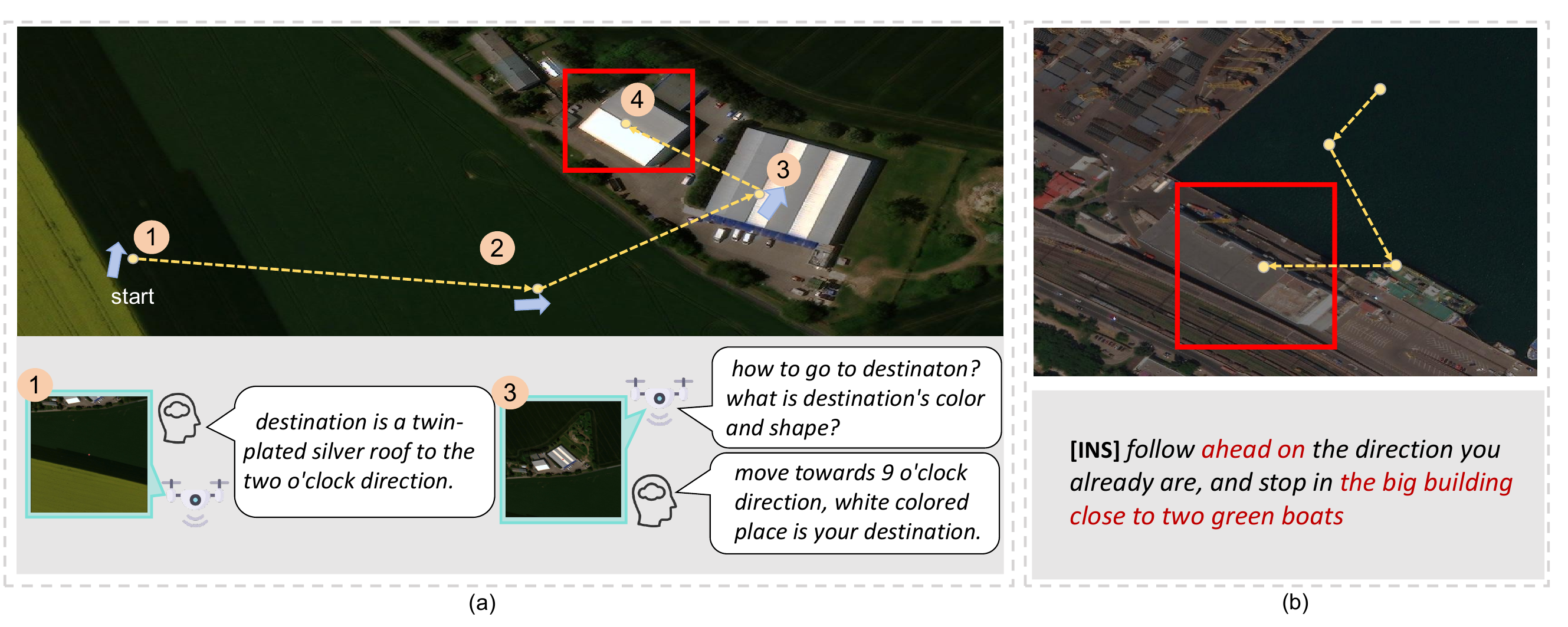}
    \caption{Visualization of navigation trajectories from our PSC-AVDN. (a) represents a two-round dialogue case, and (b) represents a single-round instruction case. The yellow dashed line indicates the navigation trajectory, while the red rectangle denotes the target area.}
    \label{fig:4}
\end{figure*}
% 需要：\usepackage{booktabs,xcolor,amssymb,array}
\newcolumntype{C}[1]{>{\centering\arraybackslash}p{#1}}

\textbf{SSM.}
As shown in Table~\ref{tab:ht_msv_gs_results}, we also perform ablation studies on the components of the SSM module.
After incorporating SVM, all three performance metrics improve noticeably.
This suggests that historical memory integration enhances spatiotemporal consistency, enabling the model to navigate more stably and accurately over extended sequences.
\begin{table}[!htbp]
\centering
\small
\setlength{\tabcolsep}{1.8pt}
\renewcommand{\arraystretch}{1.10}

{\setlength{\aboverulesep}{0pt}%
 \setlength{\belowrulesep}{0pt}%
 % 删除 Index 列后，列宽重新调整
\begin{tabular}{@{\hspace{1.8pt}} C{1.2cm}C{1.2cm}C{1.4cm}|C{1.0cm}C{1.0cm}C{1.0cm} @{\hspace{1.8pt}}}

 \toprule
 \rowcolor{gray!10}
 \textbf{SVM} & \textbf{MVO} & \textbf{SGM} & \textbf{SPL} & \textbf{SR} & \textbf{GP} \\
 \midrule
  &  &  & 16.3 & 19.3 & 35.7 \\
 \checkmark &  &  & 16.5 & 20.4 & 36.6 \\
 \checkmark & \checkmark &  & 16.6 & 21.1 & 38.3 \\
 \checkmark &  & \checkmark & 16.7 & 21.2 & 37.0 \\
  & \checkmark & \checkmark & 16.6 & 20.4 & 37.5 \\
 \checkmark & \checkmark & \checkmark & \textbf{17.8} & \textbf{22.6} & \textbf{39.2} \\
 \bottomrule
 \end{tabular}%
}
\caption{Ablation results of SVM (Spatial Visual Memory), MVO (Multi-scale Visual Observation), and SGM (Structured Geometric Memory) components in our SSM module.}
\label{tab:ht_msv_gs_results}
\end{table}
Subsequently, incorporating MVO leads to further performance gains, indicating that multi-scale visual inputs improve the model's ability to perceive both local and global visual cues.
Finally, the addition of SGM results in the best overall performance.
This shows that generating a reference grid map enhances spatial reasoning and contributes to improved navigation accuracy.
Additional ablation studies on SVM+SGM and MVO+SGM combinations further validate the individual effectiveness of each module.

\newcolumntype{C}[1]{>{\centering\arraybackslash}p{#1}}

\begin{table}[!htbp]
\centering
\small
\setlength{\tabcolsep}{1.8pt}
\renewcommand{\arraystretch}{1.10}

{\setlength{\aboverulesep}{0pt}%
 \setlength{\belowrulesep}{0pt}%
\begin{tabular}{@{\hspace{1.8pt}} C{1.6cm} | C{1.0cm} C{1.0cm} C{1.0cm} @{\hspace{1.8pt}}}
 \toprule
 \rowcolor{gray!10}
 \textbf{Grid size} & \textbf{SPL} & \textbf{SR} & \textbf{GP} \\
 \midrule
 \(3\times3\)   & 16.6 & 20.7 & 35.0 \\
 \(5\times5\)   & \textbf{17.8} & \textbf{22.6} & \textbf{39.2} \\
 \(7\times7\)   & 17.0 & 20.9 & 38.9 \\
 \(10\times10\) & 16.0 & 20.0 & 38.3 \\
 \bottomrule
 \end{tabular}%
}
\caption{Ablation results of different grid sizes in the Reference Grid Map of our SSM module.}
\label{tab:grid_size_val_unseen}
\end{table}

\textbf{Gird size for reference map.}
We further evaluate the effects of different grid sizes in the Reference Grid Map in Table~\ref{tab:grid_size_val_unseen}).
The results indicate that a $5 \times 5$ grid achieves the best performance, while both larger and smaller grids lead to clear degradation.
Small grids make it difficult to distinguish fine-grained targets accurately, whereas large grids produce sparse cell-level semantics, which introduces noise and increases the burden of CoT reasoning.

\newcolumntype{C}[1]{>{\centering\arraybackslash}p{#1}}

\begin{table}[!htbp]
\centering
\small
\setlength{\tabcolsep}{1.8pt}
\renewcommand{\arraystretch}{1.10}
{\setlength{\aboverulesep}{0pt}%
 \setlength{\belowrulesep}{0pt}%
\begin{tabular}{@{\hspace{1.8pt}} C{2.0cm} | C{1.0cm} C{1.0cm} C{1.0cm} @{\hspace{1.8pt}}}
 \toprule
 \rowcolor{gray!10}
 \textbf{Scaling factor} & \textbf{SPL} & \textbf{SR} & \textbf{GP} \\
 \midrule
 (2, 4, 6) & 17.5 & 21.4 & 37.9 \\
 (3, 5, 7) & \textbf{17.8} & \textbf{22.6} & \textbf{39.2} \\
 (4, 6, 8) & 16.0 & 19.7 & 36.7 \\
 \bottomrule
 \end{tabular}%
}
\caption{Ablation results across different scaling factor combinations in the Multi-Scale Crop of our SSM module.}
\label{tab:scaling_factor}
\end{table}
\textbf{Scaling factor.}
To investigate the effect of different scaling factor combinations in Multi-Scale Crop, we conduct an ablation study by varying the set of scales used for image resampling.
As shown in Table \ref{tab:scaling_factor}, the combination of ($3$, $5$, $7$) achieves the best performance across all metrics.
We attribute this improvement to a balanced integration of global and local visual cues, where moderate scale diversity enables the model to capture overall scene structures while still preserving fine-grained details.
In contrast, smaller scales ($2$, $4$, $6$) focus excessively on local patterns and lose global context, whereas larger scales ($4$, $6$, $8$) reduce inter-scale diversity and introduce redundant visual information, leading to performance degradation.
These results confirm that choosing appropriate scaling factors is critical for obtaining more effective multi-scale visual information.

\subsection{Visualization Results}
We selected two representative cases to visualize navigation trajectories. 
Fig.~\ref{fig:4}(a) illustrates a two-round dialogue scenario, while (b) shows a single-round instruction scenario. 
In the multi-round case, the UAV starts from the initial position and receives the instruction. 
When it reaches position $3$, it enters the second dialogue round to clarify the target area, and finally arrives at position $4$. 
In the single-round instruction case, the UAV adjusts its flight path based on the instruction and reaches the target area. More visualizations and failure cases are provided in Appendices~8 and~9.

%% file: sec/5_conclusion.tex
\section{Conclusion}
\label{sec:Conclusion}
In this paper, we present PSC-AVDN, a training-free framework for Aerial Vision-and-Dialog Navigation that integrates a three-stage Parsing-Search-Confirmation reasoning pipeline with a Structured Spatial Memory (SSM) module. The parsing stage converts ambiguous instructions into stable geometric cues, Search-CoT conducts stepwise high-altitude target exploration, and Confirmation-CoT performs fine-grained verification to resolve visual ambiguity and confirm the final target. 
Meanwhile, SSM integrates multi-scale visual observation, spatial visual memory, and structured geometric memory to provide global spatial context and long-horizon consistency.
Extensive experiments show that PSC-AVDN sets new state-of-the-art performance in the training-free setting, matching or surpassing several finetuned methods. We believe this framework offers a principled way to combine explicit CoT-style reasoning with structured spatial memory for scalable and generalizable aerial embodied navigation in the future.

\section{Acknowledgments}
This research was supported in part by the National Natural Science Foundation of China (Grant No. 62502142, 62572166, 62302140, 62472139, 62461160308, U23B2010), the Natural Science Foundation of Anhui Province (Grant No. 2508085QF226), the Fundamental Research Funds for the Central Universities (Grant No. JZ2025HGTA0161), the National Key R\&D Program of China (2022ZD0115502), the Ningbo Science and Technology Innovation 2025 Major Project (2025Z034), and the ``Pioneer'' and ``Leading Goose'' R\&D Program of Zhejiang (2024C01161).
The computation is completed on the HPC Platform of Hefei University of Technology.